\documentclass[sigconf]{acmart}

\usepackage{xcolor}
\definecolor{mycolor}{RGB}{127,185,87}

\usepackage{multirow}
\usepackage[normalem]{ulem}
\useunder{\uline}{\ul}{}
\usepackage{enumitem}
\usepackage{epigraph} 
\usepackage{balance}

\AtBeginDocument{%
  }

\copyrightyear{2025}
\acmYear{2025}
\setcopyright{acmlicensed}
\acmConference[MM '25]{Proceedings of the 33rd
ACM International Conference on Multimedia}{October 27--31, 2025}{Dublin,
Ireland}
\acmBooktitle{Proceedings of the 33rd ACM International Conference on
Multimedia (MM '25), October 27--31, 2025, Dublin, Ireland}
\acmDOI{10.1145/3746027.3755605}
\acmISBN{979-8-4007-2035-2/2025/10}

\begin{document}

\title{Hardness-Aware Dynamic Curriculum Learning for Robust Multimodal Emotion Recognition with Missing Modalities}





\author{Rui Liu}
\authornote{Rui Liu is the corresponding author.}
\email{liurui_imu@163.com}
\affiliation{%
  \institution{Inner Mongolia University}
  \city{Hohhot}
  \country{China}
}

\author{Haolin Zuo}
\email{22209010@mail.imu.edu.cn}
\affiliation{%
  \institution{Inner Mongolia University}
  \city{Hohhot}
  \country{China}
}

\author{Zheng Lian}
\email{lianzheng2016@ia.ac.cn}
\affiliation{%
  \institution{Institute of Automation, Chinese Academy of Sciences}
  \city{Beijing}
  \country{China}
}

\author{Hongyu Yuan}
\email{yuanhongyu_1997@163.com}
\affiliation{%
  \institution{Inner Mongolia University}
  \city{Hohhot}
  \country{China}
}

\author{Qi Fan}
\email{fanqi1203@foxmail.com}
\affiliation{%
  \institution{Inner Mongolia University}
  \city{Hohhot}
  \country{China}
}

\renewcommand{\shortauthors}{Rui Liu, Haolin Zuo, Zheng Lian, Hongyu Yuan, and Qi Fan}

\begin{abstract}

Missing modalities have recently emerged as a critical research direction in multimodal emotion recognition (MER). Conventional approaches typically address this issue through missing modality reconstruction. However, these methods fail to account for variations in reconstruction difficulty across different samples, consequently limiting the model's ability to handle hard samples effectively. To overcome this limitation, we propose a novel Hardness-Aware Dynamic Curriculum Learning framework, termed \textbf{HARDY-MER}. Our framework operates in two key stages: first, it estimates the hardness level of each sample, and second, it strategically emphasizes hard samples during training to enhance model performance on these challenging instances. Specifically, we first introduce a \textit{Multi-view Hardness Evaluation} mechanism that quantifies reconstruction difficulty by considering both Direct Hardness (modality reconstruction errors) and Indirect Hardness (cross-modal mutual information). Meanwhile, we introduce a \textit{Retrieval-based Dynamic Curriculum Learning} strategy that dynamically adjusts the training curriculum by retrieving samples with similar semantic information and balancing the learning focus between easy and hard instances. Extensive experiments on benchmark datasets demonstrate that HARDY-MER consistently outperforms existing methods in missing-modality scenarios. Our code will be made publicly available at \url{https://github.com/AI-S2-Lab/HARDY-MER}.

\end{abstract}

\ccsdesc[500]{Human-centered computing~Human computer interaction (HCI)}

\keywords{Multimodal Emotion Recognition, Missing Modalities Learning, Dynamic Curriculum Learning, Hardness-Aware Retrieval Augmented}


\maketitle

 \section{Introduction}
\label{sec: Introduction}

Multimodal Emotion Recognition (MER) with missing modalities has emerged as a critical research direction in affective computing \cite{tellamekala2023cold, yuan2023noise, lian2023gcnet, vazquez2023accommodating, luo2023multimodal, wang2024incomplete, sun2025enhancing}. In real-world scenarios, missing modalities frequently occur due to device failures \cite{zhao2021missing, vazquez2023accommodating, song2022multimodal, song2025diffcl}, asynchronous signals \cite{shen2020memor, lin2023enhancing}, or low-quality inputs (e.g., degraded videos) \cite{yuan2021transformer, wang2022m2r2}. However, most existing models are trained on complete-modality data, leading to poor performance under missing conditions and limiting their robustness in practical applications.

To mitigate these challenges, researchers have explored various methods and achieved significant progress \cite{cai2018deep, du2018semi, yuan2021transformer, zhao2021missing, tang2021ctfn, lian2023gcnet}. Among these efforts, mainstream methods focus on reconstructing missing modalities using available modalities \cite{zhao2021missing, zuo2023exploiting, yuan2023noise, liu2024contrastive}.
For instance, Zhao et al. \cite{zhao2021missing} proposed an imagination network to recover missing modalities and to learn the joint representation. Yuan et al. \cite{yuan2023noise} employed a diffusion model framework, leveraging available modalities to guide the generation of missing modalities and integrating the generated results with available information as a joint representation. Liu et al. \cite{liu2024contrastive} further improved the reconstruction process using modality-invariant features to strengthen model robustness under incomplete inputs.

\begin{figure}
    \centering
    \includegraphics[width=0.9\linewidth]{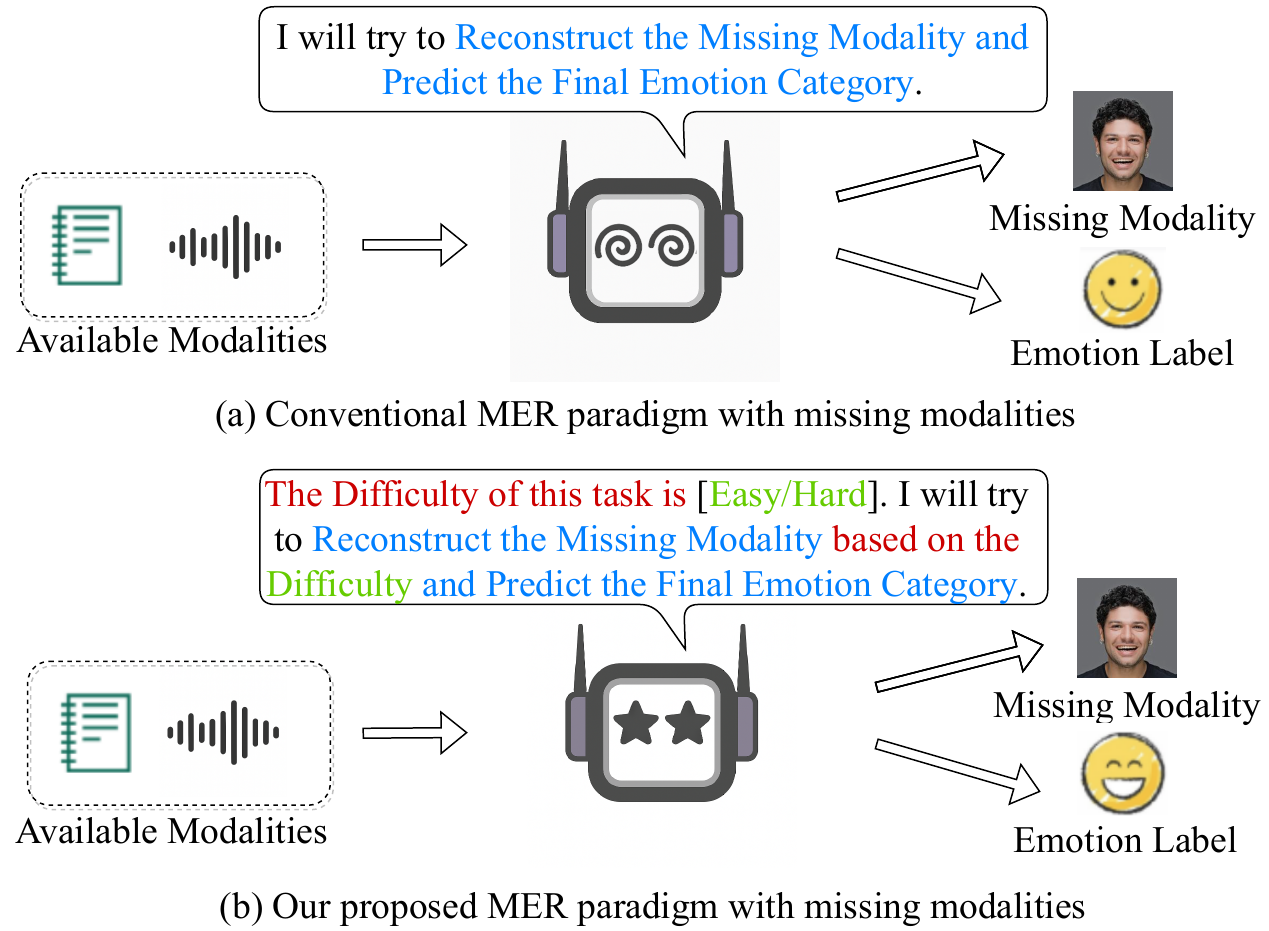}
    \caption{Comparison between conventional paradigms for emotion recognition under missing modalities and our proposed HARDY-MER.
    (a) Conventional methods attempt to reconstruct the missing modality and predict emotions without considering reconstruction difficulty, which may lead to suboptimal handling of hard samples.
    (b) Our proposed HARDY-MER first estimates the sample-specific difficulty, then allocates more attention to hard samples based on the estimated difficulty, thereby enhancing the model’s robustness in emotion recognition for challenging instances.
    }
    \label{fig: samples}
\end{figure}

Despite recent advances, a critical limitation remains: conventional methods treat all training samples equally, overlooking the varying difficulty of reconstructing missing modalities across different instances, as illustrated in Fig.~\ref{fig: samples}(a).
This homogeneous training strategy fails to acknowledge that certain samples are inherently harder to reconstruct due to factors such as semantic ambiguity, low signal quality, or strong inter-modal dependencies.
Consequently, models tend to overfit on easy samples while underexploiting harder ones, ultimately limiting their ability to generalize and adapt to complex real-world scenarios\cite{zhou2023learning}.

To address this limitation, we draw inspiration from educational psychology~\cite{brown2014make}, where students often perform more exercises on harder concepts to enhance their understanding. Motivated by this strategy, we retrieve semantically similar examples for hard samples and integrate them into training, thereby encouraging the model to focus more on these challenging instances. We call this novel framework \textbf{Har}dness-Aware \textbf{Dy}namic Curriculum Learning, termed \textbf{HARDY-MER}. To achieve this, we mainly need two key functions: hardness measurement and hardness-aware training. First, we develop a \textbf{Multi-view Hardness Evaluation} mechanism that quantifies hardness based on two criteria: \emph{direct hardness}, measured by reconstruction errors across modalities, and \emph{indirect hardness}, assessed through mutual information between modalities. This dual-perspective evaluation enables a more comprehensive and accurate hardness assessment. Second, to prioritize harder examples during training, we propose a \textbf{Retrieval-based Dynamic Curriculum Learning} strategy. Specifically, we design a retrieval mechanism that fuses \textit{local similarities} across available modalities into a unified \textit{global similarity} score of the sample, which is then used to identify the most relevant candidate samples. The number of retrieved samples is then dynamically adjusted based on estimated hardness, allocating more training resources to harder samples while reducing emphasis on easier ones. The main contributions of this paper are as follows:


\begin{itemize}

\item 
We propose a novel \textbf{Multi-view Hardness Evaluation} mechanism that jointly models direct and indirect hardness to facilitate comprehensive, modality-sensitive training hardness estimation.

\item 
We introduce a \textbf{Retrieval-based Dynamic Curriculum Learning} strategy that dynamically retrieves semantically relevant samples based on estimated hardness and adaptively adjusts their number to balance learning between easy and hard instances, therefore enhancing model robustness under missing modality conditions.

\item 
Extensive experiments on IEMOCAP and CMU-MOSEI across six missing modality settings demonstrate the superiority of our method over existing baselines, achieving new state-of-the-art results in per-condition metrics.

\end{itemize}

\vspace{-2mm}
\section{Related Work}
\label{sec: Related Work}

\subsection{Hard Sample Mining}
\label{subsec: Hard Sample Mining}
Hard sample mining is a popular technique for enhancing a model's discriminative ability, widely applied in tasks such as face recognition \cite{schroff2015facenet}, object detection \cite{shrivastava2016training, wang2018towards}, speech separation \cite{wang2022mining}, and masked image/audio reconstruction \cite{wang2023hard, seth2024eh}, etc.
Related studies have shown that hard samples frequently serve as model performance bottlenecks \cite{li2023focus, tang2023multiple, wu2024hgmd}, and targeting these challenging instances can produce significant performance improvements \cite{wang2023hard, seth2024eh, lin2024rho}.
For example, Li et al. \cite{li2019deep} utilized attention scores to pinpoint important instances from false negative bags, which were then used as hard negative instances to create hard bags, ultimately enhancing classification performance.
Wang et al. \cite{wang2023hard} measured the reconstruction hardness of samples based on reconstruction error and performed masked reconstruction on image patches with higher reconstruction hardness to improve the model's ability to reconstruct masked images, thereby enhancing the robustness of visual representation learning.
Tang et al. \cite{tang2023multiple} proposed a teacher-student framework with consistency constraints for multi-instance classification tasks.
In this approach, the teacher model implicitly mines hard instances based on attention scores, which are then used to train the student model, enabling the student to learn better discriminative boundaries.

However, when applied to multimodal tasks, traditional hard example mining methods face the following limitations:
1) Even when a modality is present in the input, its reconstruction hardness can still indicate whether its semantic information is redundant or complementary to other modalities \cite{yuan2021transformer, lian2023gcnet}.
A high reconstruction error for an observed modality suggests that the information it carries cannot be easily inferred from the others, thus making the sample intrinsically difficult for the model to learn.
2) Although single metrics such as reconstruction loss \cite{wang2023hard, seth2024eh} or attention scores \cite{tang2023multiple} can be used to estimate sample hardness, they may not sufficiently capture the complexity of multimodal learning.
In particular, these approaches often overlook the importance of cross-modal consistency \cite{hazarika2020misa, liu2024contrastive}.
Samples that are easy to reconstruct in individual modalities may still pose learning challenges when cross-modal consistency is weak \cite{lin2021completer}.

To overcome the limitations described above, we propose a composite metric to comprehensively evaluate the learning hardness of multimodal samples.
Specifically, our metric consists of two components: direct hardness, which intuitively reflects the sample’s difficulty by assessing the reconstruction error of each modality; and indirect hardness, which measures the level of mutual information between different modalities, capturing the sample’s challenge from the perspective of cross-modal consistency.
By combining these two perspectives, the proposed metric provides a more comprehensive and reliable estimation of the hardness of the sample.
This serves as a foundation for the subsequent retrieval of samples and the construction of the curriculum.

\vspace{-2mm}
\subsection{Retrieve Augmented Generation}

Retrieval-augmented generation (RAG) is a hybrid approach that integrates information retrieval with generative models, aiming to enhance the quality and accuracy of generation tasks.
This method equips pre-trained generative models with the ability to incorporate non-parametric memory, enabling them to effectively leverage external knowledge \cite{lewis2020retrieval}.
In NLP tasks, RAG improves the quality of text generation by retrieving relevant documents \cite{lewis2020retrieval, guu2020retrieval, izacard2020leveraging, borgeaud2022improving, wu2024retrieval}.
For example, Borgeaud et al. \cite{borgeaud2022improving} proposed the Retrieval-Enhanced Transformer that enhances auto-regressive language models by conditioning on document chunks retrieved from a large corpus, based on local similarity with preceding tokens.
Moreover, RAG has also been applied to dialogue generation tasks \cite{liu2025retrieval}, where it is used to generate expressive speech that aligns with conversational styles.
The traditional RAG method mainly focuses on directly incorporating the retrieved information into the generation process to improve output quality.
In contrast, our approach enhances the training process by using retrieval techniques to find similar samples for challenging instances.
Additionally, this is the first work to apply RAG technology to multimodal emotion recognition with missing modality.

\vspace{-5pt}
\subsection{Curriculum Learning}
\label{subsec: Curriculum Learning} 


Curriculum learning (CL) is a training strategy inspired by the structurally sequential learning approach in human education \cite{graves2016hybrid, wu2022training, zhao2023m2df}.
Its core idea is to “start small,” using an easier subset of data to train the model, and then gradually incorporating more challenging data until the entire training dataset is covered \cite{bengio2006greedy, wei2016stc, wang2021survey, zhou2023learning}.
Typically, curriculum learning utilizes a predefined \cite{zhou2023clcl, zhou2023learning, zhao2023m2df} or automatically learned \cite{jiang2014easy, jiang2014self, jiang2015self, wang2023hard, seth2024eh} difficulty predictor to distinguish between easier and harder samples, followed by a training scheduler that determines how to introduce the more challenging samples into the training process.
CL not only accelerates the training process \cite{jiang2014self, platanios2019competence} but also enhances the model’s generalization capability \cite{wang2023efficienttrain}.
Recent extensive research has demonstrated the remarkable effectiveness of curriculum learning in fields such as computer vision \cite{wu2022training, wang2023efficienttrain}, human-object interaction detection \cite{zhou2023learning}, acoustic representation learning \cite{seth2024eh}, etc.
However, influenced by the ``easy-to-hard'' training paradigm, traditional curriculum learning often prioritizes easy samples while inadequately addressing hard samples. Our method differs from these approaches in several notable aspects: 
1) We innovatively integrate retrieval augmentation into curriculum learning, enabling semantic-aware instance expansion to enhance training sample diversity; 
2) During retrieval, we incorporate sample difficulty signals to provide more semantically similar instances for challenging samples, therefore strengthening the model’s capability to learn from hard cases.
To the best of our knowledge, this represents the first approach that systematically unifies retrieval techniques with curriculum learning.

\section{Methodology}
\label{sec: Methodology}

\begin{figure*}[]
    \centering
    \includegraphics[width=0.9\linewidth]{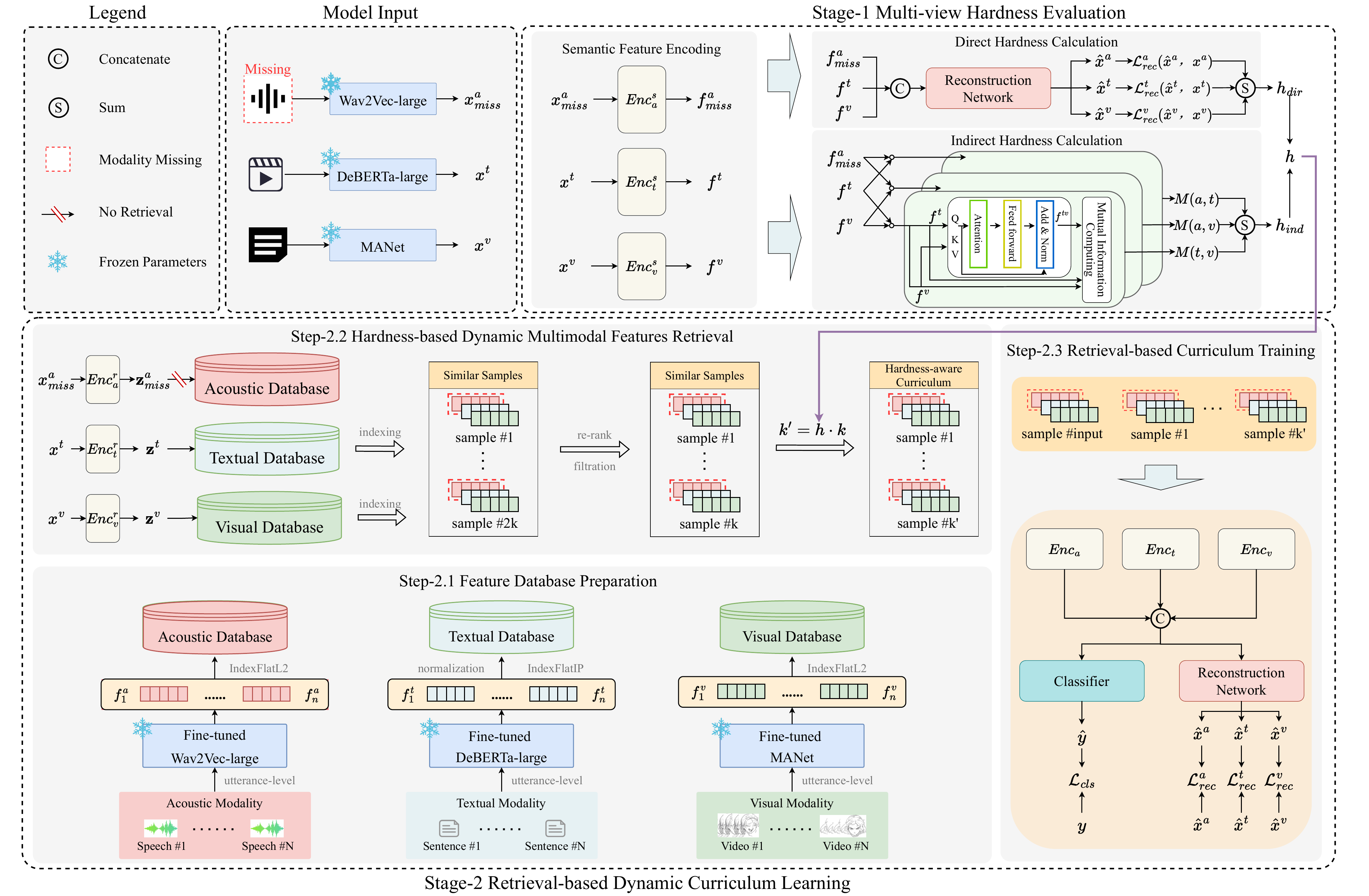}
    \caption{The overview of HARDY-MER consists of Multi-view Hardness Evaluation, Feature Database Preparation, Hardness-based Dynamic Multimodal Features Retrieval, and Retrieval-based Curriculum Training.}
    \label{fig: overview}
\end{figure*}

\subsection{Overview}
\label{subsec: Overview}
As shown in Fig. \ref{fig: overview}, the proposed HARDY-MER includes two main components: 
\textbf{1) Multi-view Hardness Evaluation} simulates the role of a teacher by assessing the hardness of input samples based on the reconstruction errors of missing modalities and the mutual information across available modalities; 
\textbf{2) Retrieval-based Dynamic Curriculum Learning} is designed to retrieve semantically similar samples, construct dynamic curricula, and train the model accordingly.
This process consists of three key steps:
\textbf{a) Feature Database Preparation}, which builds a multimodal feature index for semantic retrieval;
\textbf{b) Hardness-based Dynamic Multimodal Feature Retrieval}, which selects the most relevant samples based on the input’s available modalities and adaptively adjusts the retrieval size according to the input’s estimated hardness, assigning more training resources to more challenging samples while allocating fewer to easier ones;
and \textbf{c) Retrieval-based Curriculum Training}, where the emotion recognition model is trained using the resulting curriculum.

\subsection{Multi-view Hardness Evaluation}
\label{subsec: Multi-view Hardness Evaluation}

To quantify the learning hardness of each training sample under missing-modality conditions, we propose a unified metric termed \textbf{multi-view hardness}, which consists of two complementary components: (1) \emph{direct hardness}, reflecting the reconstruction error of the modalities, and (2) \emph{indirect hardness}, measuring the level of mutual information between different modalities.
Stage 1 of Fig.~\ref{fig: overview} illustrates the overall computation process of this multi-view hardness evaluation.
In what follows, we detail the formulation of both metrics and describe the training strategy for the hardness evaluation module.

\subsubsection{Semantic Representation Extraction}

Given a multimodal input sample $(x^a_{\text{miss}}, x^t, x^v)$, we first extract modality semantic features using a \textit{Semantic Feature Encoding} module.
This module employs three Transformer-based encoders~\cite{xu2024leveraging} to produce representations $(f^a_{\text{miss}}, f^t, f^v)$, where the subscript “miss” indicates that the corresponding modality is absent.
Following prior work~\cite{zhao2021missing, lian2023gcnet, liu2024contrastive}, we represent the missing modality using a zero vector.
These semantic representations are used to compute both direct and indirect hardness scores.

\subsubsection{Hardness Metric Computation}

\paragraph{\textbf{Direct Hardness}.}
To estimate direct hardness, we concatenate the semantic features from the three modalities and pass them through a linear reconstruction network to recover each modality:
\begin{equation}
\hat{x}^m = W_m \cdot [f^a_{\text{miss}}; f^t; f^v] + b_m, \quad m \in \{a, t, v\},
\end{equation}
where $\hat{x}^m$ denotes the reconstructed feature of modality $m$, and $a, t, v$ denote acoustic, textual, and visual modalities, respectively. 
$W_m$, $b_m$ are trainable parameters.
$[\cdot ; \cdot]$ denotes feature concatenation across modalities.
We adopt the Mean Squared Error (MSE) loss \cite{wang2023hard, lin2024rho} to measure the reconstruction quality of each modality:
\begin{equation}
h^m_{\text{dir}} = \mathcal{L}^m_{\text{rec}}(\hat{x}^m, x^m),
\end{equation}
and define the overall direct hardness as:
\begin{equation}
h_{\text{dir}} = h^a_{\text{dir}} + h^t_{\text{dir}} + h^v_{\text{dir}}.
\end{equation}
Note that the reconstruction loss $\mathcal{L}^m_{\text{rec}}$ is used only for hardness estimation and does not participate in gradient backpropagation.

\paragraph{\textbf{Indirect Hardness}.}


In the Indirect Hardness Calculation module, we compute the mutual information (MI) between each pair of modalities using their semantic features $(f^a_{\text{miss}}, f^t, f^v)$.
Following the standard definition of mutual information:
\begin{equation}
I(X; Y) = H(X) + H(Y) - H(X, Y),
\end{equation}
where $H(\cdot)$ denotes entropy.
However, estimating the joint entropy $H(X, Y)$ directly in high-dimensional feature spaces is notoriously challenging.
To address this, we adopt the strategy proposed by Huang et al. \cite{huang2023dominant}, which approximates the joint distribution via fusion features.
Specifically, for a given modality pair $f^p$ and $f^q$, we first apply a cross-attention mechanism to fuse them, treating one modality as the query and the other as the key-value input:
\begin{equation}
f^{p \rightarrow q} = \text{CrossAttn}(f^p, f^q, f^q).
\end{equation}
To ensure symmetric information capture, we swap the query and key-value roles and repeat the operation:
\begin{equation}
f^{q \rightarrow p} = \text{CrossAttn}(f^q, f^p, f^p).
\end{equation}
The final joint representation is then obtained by summing the two fused outputs:
\begin{equation}
f^{p,q} = f^{p \rightarrow q} + f^{q \rightarrow p}.
\end{equation}
We then estimate the entropy of each individual modality, $H(f^p)$ and $H(f^q)$, as well as the entropy of the fused representation $H(f^{p,q})$. The mutual information between $p$ and $q$ is calculated as follows:
\begin{equation}
I(p;q) = H(f^p) + H(f^q) - H(f^{p,q}).
\end{equation}
Finally, we define the indirect hardness $h_{\text{ind}}$ as the sum of the mutual information between the modalities:
\begin{equation}
h_{\text{ind}} = I(a;t) + I(a;v) + I(t;v).
\end{equation}

\paragraph{\textbf{Unified Hardness Score}.}

We combine direct and indirect hardness into a final unified score using a scaled logistic function:
\begin{equation}
h = (1 + \exp(-\beta \cdot (\alpha_1 \cdot h_{\text{dir}} + \alpha_2 \cdot h_{\text{ind}})))^{-1}.
\label{eq: hardness}
\end{equation}
where $\alpha_1$ and $\alpha_2$ are weighting factors that balance the contributions of direct and indirect hardness, and $\beta$ is a scaling coefficient that controls the sharpness of the transition. 
This formulation normalizes the hardness score to the range $(0, 1)$, enabling a smooth and differentiable measure that reflects the overall learning hardness of a sample.

\subsubsection{Hardness Module Training}

To ensure the reliability of the estimated hardness scores, we adopt a two-stage training strategy for the Multi-view Hardness Evaluation module. Indirect and direct hardness are trained separately on full and missing modality data, respectively.

\textbf{Stage 1: Indirect Hardness Training.}
We first train the semantic encoders and Indirect Hardness Calculation components on complete multimodal samples.
The training objective in this stage includes two parts:  
1) the supervised classification loss based on modality features $f^m$, encouraging the encoders to capture sentiment-discriminative information:
\begin{equation}
    \hat{y}^m = \text{CLS}_m(f^m), \quad m \in (a,t,v),
\end{equation}
\begin{equation}
    \mathcal{L}^1_{\text{cls}} = \sum_{m \in \{a, t, v\}} \text{CE}(y, \hat{y}^m),
\end{equation}
where $\text{CLS}_m$ denotes a classification head for corresponding modality based on the fully-connected layer, $\hat{y}^m$ is the predicted emotion class for modality $m$, and $y$ is the ground truth.
$\text{CE}(\cdot, \cdot)$ represents the standard cross-entropy loss function.
2) the mutual information regularization loss:
\begin{equation}
\mathcal{L}_{\text{MI}} = - h_{\text{ind}},
\end{equation}
which ensures the reliability of mutual information estimation by encouraging the module to capture consistent cross-modal information.
The total loss in the stage is: $\mathcal{L}^1_{total} = \mathcal{L}^1_{cls} + \mathcal{L}_{MI}$.

\textbf{Stage 2: Direct Hardness Training.}
We further fine-tune the Semantic Feature Encoder components and jointly train the Direct Hardness Calculation module using samples with missing modalities.  
Given the semantic features extracted from the available modalities, we first perform emotion classification using the concatenated features:
\begin{equation}
    \hat{y} = \text{CLS}([f^a_{\text{miss}}; f^t; f^v]),
\end{equation}
\begin{equation}
    \mathcal{L}^2_{\text{cls}} = \text{CE}(y, \hat{y}).
\end{equation}
In parallel, we compute the direct hardness based on modality reconstruction error:
\begin{equation}
\mathcal{L}_{\text{rec}} = h_{\text{dir}}.
\end{equation}
The total loss for this stage is defined as:
$\mathcal{L}^2_{\text{total}} = \mathcal{L}^2_{\text{cls}} + \mathcal{L}_{\text{rec}}$.

After the two-stage training, the parameters of the Multi-view Hardness Evaluation module are frozen and used throughout the rest of the framework.


\subsection{Retrieval-based Dynamic Curriculum Learning}
\label{subsec: Retrieval-based Dynamic Curriculum Learning}
Stage 2 in Fig.~\ref{fig: overview} illustrates the structure of the \textbf{Retrieval-Based Dynamic Curriculum Learning} module, which consists of three steps: \textit{Feature Database Preparation}, \textit{Hardness-based Dynamic Multimodal Feature Retrieval}, and \textit{Retrieval-based Curriculum Training}.

\subsubsection{Feature Database Preparation}
\label{subsubsec: Feature Database Preparation}
As shown in Step 2-1 of Fig.~\ref{fig: overview}, to enhance the semantic consistency between stored and retrieved features during training, we employ a fine-tuned pre-trained model for feature extraction.
Furthermore, we employ distinct index construction strategies for different modalities to optimize retrieval performance.

\textbf{Features Preparation: }
We fine-tune pre-trained models via the emotion classification task to extract emotion features.
Specifically, we use DeBERTa-large\footnote{\url{https://huggingface.co/microsoft/deberta-large}}, Wav2Vec-large\footnote{\url{https://github.com/pytorch/fairseq/tree/main/examples/wav2vec}}, and MANet\footnote{\url{https://github.com/zengqunzhao/MA-Net}} as frozen backbones for textual, acoustic, and visual modalities, respectively.
Two trainable linear layers are appended to each backbone, and the output of the final layer is used as the semantic feature for retrieval.
A classification head is trained with cross-entropy loss to guide the feature extraction toward sentiment-relevant representations.

\textbf{Database Construction:}  
We utilize the FAISS (Facebook AI Similarity Search) library to construct modality semantic feature databases, applying tailored similarity metrics based on the characteristics of each modality.
For textual features, we normalize all vectors and use \textit{IndexFlatIP}\footnote{\textit{IndexFlatIP} and \textit{IndexFlatL2} are two commonly used exact search index types in the FAISS library, corresponding to inner product and Euclidean distance, respectively.} to implement cosine similarity.
For acoustic and visual features, we adopt \textit{IndexFlatL2} to perform Euclidean distance-based retrieval.
This process results in three separate databases for text, audio, and visual modalities.
The effectiveness of this configuration is validated in Sec.~\ref{subsec: visualization analysis}, where we compare alternative index strategies and demonstrate the superiority of our method.

\subsubsection{Hardness-based Dynamic Multimodal Features Retrieval}
\label{subsubsec: Multimodal Features Retrieval}

This module is illustrated in Step 2.2 of Fig.~\ref{fig: overview}. 
Given an input sample $(x^a_{\text{miss}}, x^t, x^v)$, we first use modality semantic encoders $\text{Enc}^r_m$ to extract high-level embeddings for each modality:
\begin{equation}
\mathbf{z}^m = \text{Enc}^r_m(x^m), \quad m \in \{a, t, v\}.
\end{equation}
For each available modality, we query its corresponding FAISS index using the embedding $\mathbf{z}^m$ to retrieve the top-$k$ most semantically similar samples, and record their indices.  
We then aggregate the indices retrieved from all available modalities and remove duplicates to construct a unified candidate set.  
Based on these indices, we retrieve the corresponding multimodal features (acoustic, textual, and visual) from the three modality feature databases.  
The features retrieved under the same index collectively form a candidate sample.

To evaluate the overall similarity between a candidate and the input sample, we compute the L2 distance between their corresponding features in each available modality of the candidate sample.  
We then take the average of these distances as the integrated similarity score.  
Finally, we rank all candidate samples in ascending order of similarity and select the top-$k$ most similar ones as the final retrieval results.

Based on the retrieval results, we further construct a hardness-aware curriculum to guide model training.  
To ensure that harder samples receive more support while easier ones receive less, we use the sample hardness score $h \in (0, 1)$ from Stage-1 to adaptively determine the number of support samples:
\begin{equation}
k' = \lceil h \cdot k \rceil,
\end{equation}
We then select the top-$k'$ entries from the retrieval results as the hardness-aware curriculum for training.

\subsubsection{Retrieval-based Curriculum Training}
\label{subsubsec: Retrieval-based Curriculum Scheduler}
As illustrated in Step 2.3 of Fig.~\ref{fig: overview}, we integrate the input sample $(x^a_{\text{miss}}, x^t, x^v)$ with its corresponding hardness-aware curriculum retrieved in Step 2.2 to train our emotion recognition model.  
The model consists of three Transformer-based modality encoders, a reconstruction network based on autoencoders, and a classification head.
To ensure that each encoder extracts robust semantic representations, we follow the previous works \cite{liu2024contrastive, xu2024leveraging} and adopt a two-stage training strategy.

\textbf{Stage 1: Pretraining with complete modality input.}  
We feed the full input $(x^a, x^t, x^v)$ into the corresponding encoders $\text{Enc}^m$ to obtain complete modality semantic features $(f^a, f^t, f^v)$.  
To supervise the representation learning of each modality, we attach three independent classification heads and perform sentiment prediction using the features from each modality separately.  
The classification heads are trained with cross-entropy loss to guide each encoder in capturing discriminative sentiment-related information.

\textbf{Stage 2: Curriculum-based model training.}  
We then train the full model using the hardness-aware curriculum generated for each input.  
Each training instance consists of the original input $(x^a_{\text{miss}}, x^t, x^v)$ followed by its retrieved support samples, ordered from high to low similarity.  
The concatenated semantic features from the three encoders are jointly used for both emotion classification and missing modality reconstruction.  
During this stage, the entire model is optimized using a combination of classification loss and reconstruction loss, which jointly encourage accurate emotion prediction and robust recovery of the missing modality.

During inference, we use the trained model to perform emotion prediction on inputs with missing modalities, without requiring dynamic curriculum retrieval.

\section{Experiment}
\label{sec: experiment}

\subsection{Datasets and Evaluation Metrics}
\label{subsec: dataset and evaluation metrics}

To validate the effectiveness of our approach, we conducted extensive experiments on two public benchmark datasets: 



\textbf{IEMOCAP}~\cite{busso2008iemocap} is a widely adopted benchmark dataset for multimodal emotion recognition. It is commonly used in prior studies for both four-class classification (i.e., \textit{Happy}, \textit{Sad}, \textit{Neutral}, \textit{Angry})~\cite{zhao2021missing, zuo2023exploiting, liu2024contrastive} and six-class classification (i.e., \textit{Happy}, \textit{Angry}, \textit{Sad}, \textit{Neutral}, \textit{Surprised}, \textit{Fearful})~\cite{majumder2019dialoguernn, mai2020modality, lian2023gcnet}. In this work, we evaluate our method under both settings to ensure a comprehensive comparison with existing approaches.

\textbf{CMU-MOSEI} is a benchmark dataset for multimodal sentiment analysis, comprising 22856 annotated video clips collected from YouTube. Each utterance is labeled with a continuous sentiment score ranging from $-3$ to $+3$, indicating its polarity and intensity. Following prior work~\cite{xu2024leveraging}, we formulate this task as binary sentiment classification by labeling utterances with scores greater than zero as \textit{positive}, and those with scores less than zero as \textit{negative}.

For the IEMOCAP dataset, we follow previous work \cite{zhao2021missing, zuo2023exploiting, liu2024contrastive, xu2024leveraging} and use weighted accuracy (WA) and unweighted accuracy (UA) as evaluation metrics. For the CMU-MOSEI dataset, we use accuracy (Acc) and F1 score as evaluation metrics \cite{xu2024leveraging}.

\begin{table*}[!htbp]
\caption{Performance comparison with state-of-the-art methods (SOTA) under six possible missing modality conditions on two benchmark datasets. “Average” refers to the average performance of the models across all six missing modality conditions. The best results in each dataset are highlighted in bold, and the second-best results are underlined. The row marked with $\Delta_{Sota}$ indicates the improvement or reduction of our method compared to the best-competing method. We perform a T-test on the Average column and $\ast$ indicates that the p-value $<$ 0.05.}
\resizebox{1\textwidth}{!}{
\begin{tabular}{c|l|cccccccccccccc}
\hline \hline
\multirow{3}{*}{Dataset}     & \multicolumn{1}{c|}{\multirow{3}{*}{model}} & \multicolumn{14}{c}{Testing Condition}                                                                                                                                                                                                                    \\ \cline{3-16} 
                             & \multicolumn{1}{c|}{}                       & \multicolumn{2}{c}{a}             & \multicolumn{2}{c}{v}             & \multicolumn{2}{c}{t}             & \multicolumn{2}{c}{at}            & \multicolumn{2}{c}{av}            & \multicolumn{2}{c}{tv}            & \multicolumn{2}{c}{Average}       \\ \cline{3-16} 
                             & \multicolumn{1}{c|}{}                       & WA              & UA              & WA              & UA              & WA              & UA              & WA              & UA              & WA              & UA              & WA              & UA              & WA              & UA              \\ \hline
\multirow{7}{*}{\begin{tabular}[c]{@{}c@{}}IEMOCAP\\ four-class\end{tabular}} & CPMNet \cite{zhang2020deep}                                      & 0.4685          & 0.5172          & 0.4495          & 0.4449          & 0.4563          & 0.4532          & 0.3481          & 0.3623          & 0.4867          & 0.4933          & 0.4562          & 0.4657          & 0.4442          & 0.4561          \\
                             & MMIN \cite{zhao2021missing}                                        & 0.5658          & 0.5900          & 0.5252          & 0.5060          & 0.6657          & 0.6802          & 0.7294          & 0.7114          & 0.6399          & 0.6343          & 0.7167          & 0.6861          & 0.6405          & 0.6347          \\
                             & GCNet \cite{lian2023gcnet}                                       & 0.6558          & 0.6876          & {\ul 0.5796}    & {\ul 0.5254}    & 0.7233          & 0.7042          & 0.7702          & 0.7687          & 0.6740          & 0.6564          & {\ul 0.7563}    & 0.7362          & 0.6932          & 0.6798          \\
                             & CIF-MMIN \cite{liu2024contrastive}                                   & 0.5753          & 0.6006          & 0.5346          & 0.5156          & 0.6722          & 0.6899          & 0.7419          & 0.7259          & 0.6499          & 0.6353          & 0.7240          & 0.6991          & 0.6497          & 0.6444          \\
                             & MoMKE \cite{xu2024leveraging}                                      & {\ul 0.6953}    & {\ul 0.7021}    & 0.5680          & 0.5203          & {\ul 0.7730}    & {\ul 0.7766}    & {\ul 0.7903}    & {\ul 0.7988}    & {\ul 0.6857}    & {\ul 0.6622}    & 0.7555          & {\ul 0.7418}    & {\ul 0.7113}    & {\ul 0.7003}    \\
                             & HARDY-MER (our)                                         & \textbf{0.7265} & \textbf{0.7387} & \textbf{0.6319} & \textbf{0.6054} & \textbf{0.8249} & \textbf{0.8269} & \textbf{0.8167} & \textbf{0.8243} & \textbf{0.7419} & \textbf{0.7450} & \textbf{0.7918} & \textbf{0.7851} & \textbf{0.7556} & \textbf{0.7542} \\
                             & $\Delta_{Sota}$                                 & $\uparrow$0.0312          & $\uparrow$0.0366          & $\uparrow$0.0523          & $\uparrow$0.0800          & $\uparrow$0.0519          & $\uparrow$0.0503          & $\uparrow$0.0264          & $\uparrow$0.0255          & $\uparrow$0.0562          & $\uparrow$0.0828          & $\uparrow$0.0355          & $\uparrow$0.0433          & $\uparrow$0.0443$^\ast$          & $\uparrow$0.0539$^\ast$          \\ \hline
\multirow{7}{*}{\begin{tabular}[c]{@{}c@{}}IEMOCAP\\ six-class\end{tabular}}  & CPMNet \cite{zhang2020deep}                                     & 0.2947          & 0.2980          & 0.2620          & 0.2495          & 0.3244          & 0.3495          & 0.3349          & 0.3394          & 0.2692          & 0.2546          & 0.3134          & 0.3043          & 0.2998          & 0.2992          \\
                             & MMIN \cite{zhao2021missing}                                       & 0.4408          & 0.4296          & 0.3574          & 0.3065          & 0.4217          & 0.3855          & 0.5195          & 0.4831          & 0.4192          & 0.3815          & 0.4749          & 0.4063          & 0.4389          & 0.3988          \\
                             & GCNet \cite{lian2023gcnet}                                      & 0.4995          & 0.4645          & {\ul 0.3978}    & {\ul 0.3497}    & 0.5648          & 0.5562          & 0.5824          & 0.5725          & 0.4757          & 0.4331          & 0.5743          & 0.5466          & 0.5158          & 0.4871          \\
                             & CIF-MMIN \cite{liu2024contrastive}                                   & 0.4496          & 0.4356          & 0.3611          & 0.3135          & 0.4340          & 0.3971          & 0.5243          & 0.4920          & 0.4254          & 0.3922          & 0.4888          & 0.4491          & 0.4472          & 0.4133          \\
                             & MoMKE \cite{xu2024leveraging}                                      & {\ul 0.5051}    & {\ul 0.4738}    & 0.3907          & 0.3451          & {\ul 0.6109}    & {\ul 0.6019}    & {\ul 0.6318}    & {\ul 0.6194}    & {\ul 0.4865}    & {\ul 0.4408}    & {\ul 0.5992}    & {\ul 0.5755}    & {\ul 0.5374}    & {\ul 0.5094}    \\
                             & HARDY-MER (our)                                         & \textbf{0.5158} & \textbf{0.4914} & \textbf{0.4302} & \textbf{0.3649} & \textbf{0.6589} & \textbf{0.6195} & \textbf{0.6518} & \textbf{0.6298} & \textbf{0.5291} & \textbf{0.4745} & \textbf{0.6166} & \textbf{0.5786} & \textbf{0.5671} & \textbf{0.5265} \\
                             & $\Delta_{Sota}$                                 & $\uparrow$0.0107          & $\uparrow$0.0176          & $\uparrow$0.0324          & $\uparrow$0.0152          & $\uparrow$0.0480          & $\uparrow$0.0176          & $\uparrow$0.0200          & $\uparrow$0.0104          & $\uparrow$0.0426          & $\uparrow$0.0337          & $\uparrow$0.0174          & $\uparrow$0.0031          & $\uparrow$0.0297$^\ast$          & $\uparrow$0.0170$^\ast$          \\ \hline
\multirow{2}{*}{Dataset}     & \multicolumn{1}{c|}{\multirow{2}{*}{model}} & \multicolumn{2}{c}{a}             & \multicolumn{2}{c}{v}             & \multicolumn{2}{c}{t}             & \multicolumn{2}{c}{at}            & \multicolumn{2}{c}{av}            & \multicolumn{2}{c}{tv}            & \multicolumn{2}{c}{Average}       \\ \cline{3-16} 
                             & \multicolumn{1}{c|}{}                       & ACC             & F1              & ACC             & F1              & ACC             & F1              & ACC             & F1              & ACC             & F1              & ACC             & F1              & ACC             & F1              \\ \hline
\multirow{7}{*}{CMUMOSEI}    & CPMNet \cite{zhang2020deep}                                     & 0.6571          & 0.6518          & 0.6123          & 0.6173          & 0.7287          & 0.7244          & 0.7265          & 0.7224          & 0.6156          & 0.6199          & 0.6629          & 0.6684          & 0.6672          & 0.6674          \\
                             & MMIN \cite{zhao2021missing}                                       & 0.5890          & 0.5950          & 0.5930          & 0.6001          & 0.8220          & 0.8240          & 0.8370          & 0.8330          & 0.6355          & 0.6191          & 0.8175          & 0.8142          & 0.7157          & 0.7142          \\
                             & GCNet \cite{lian2023gcnet}                                      & 0.7204          & 0.7034          & {\ul 0.6808}          & {\ul 0.6725}          & 0.8426          & 0.8417          & 0.8510          & 0.8510          & 0.7149          & 0.6996          & 0.8474          & 0.8454          & 0.7762          & 0.7689          \\
                             & CIF-MMIN \cite{liu2024contrastive}                                   & 0.6387          & 0.6460          & 0.6196          & 0.6266          & 0.8353          & 0.8304          & 0.8401          & 0.8347          & 0.6468          & 0.6208          & 0.8250          & 0.8194          & 0.7343          & 0.7297          \\
                            & MoMKE \cite{xu2024leveraging}                                         & {\ul 0.7256}          & {\ul 0.7103}          & 0.6450          & 0.6346          & {\ul 0.8610}          & {\ul 0.8603}          & \textbf{0.8632}          & \textbf{0.8629}          & {\ul 0.7237}          & {\ul 0.7207}          & \textbf{0.8690}          & \textbf{0.8691}          & {\ul 0.7813}          & {\ul 0.7763}          \\
                            & HARDY-MER (our)                                           & \textbf{0.7482}          & \textbf{0.7411}          & \textbf{0.6935}          & \textbf{0.6750}          & \textbf{0.8720}          & \textbf{0.8713}          & {\ul 0.8542}          & {\ul 0.8501}          & \textbf{0.7482}          & \textbf{0.7411}          & {\ul 0.8572}          & {\ul 0.8539}          & \textbf{0.7956}          & \textbf{0.7888}          \\
                            & $\Delta_{Sota}$                             & $\uparrow$0.0226          & $\uparrow$0.0308          & $\uparrow$0.0127          & $\uparrow$0.0025          & $\uparrow$0.0110          & $\uparrow$0.0110          & $\downarrow$-0.0090         & $\downarrow$-0.0128         & $\uparrow$0.0245          & $\uparrow$0.0204          & $\downarrow$-0.0118         & $\downarrow$-0.0152         & $\uparrow$0.0143$^\ast$          & $\uparrow$0.0124$^\ast$          \\ \hline \hline
\end{tabular}
}
\label{tab: main results}
\end{table*}

\begin{table*}[!htbp]
\caption{The results of the ablation experiments under six missing conditions. We report the weighted accuracy (WA) and unweighted accuracy (UA) of these experiments on the IEMOCAP four-class task.}
\label{tab: ablation study}
\resizebox{1\textwidth}{!}{
\begin{tabular}{l|cccccccccccccc}
\hline \hline
\multicolumn{1}{c|}{\multirow{3}{*}{model}} & \multicolumn{14}{c}{Testing Condition}                                                                                                                                                                                                                   \\ \cline{2-15} 
\multicolumn{1}{c|}{}                       & \multicolumn{2}{c}{a}             & \multicolumn{2}{c}{v}             & \multicolumn{2}{c}{t}             & \multicolumn{2}{c}{at}            & \multicolumn{2}{c}{av}           & \multicolumn{2}{c}{tv}            & \multicolumn{2}{c}{Average}       \\ \cline{2-15} 
\multicolumn{1}{c|}{}                       & WA              & UA              & WA              & UA              & WA              & UA              & WA              & UA              & WA              & UA             & WA              & UA              & WA              & UA              \\ \hline
HARDY-MER (our)                                   & \textbf{0.7265} & \textbf{0.7387} & \textbf{0.6319} & \textbf{0.6054} & \textbf{0.8249} & \textbf{0.8269} & \textbf{0.8167} & \textbf{0.8243} & \textbf{0.7419} & \textbf{0.745} & \textbf{0.7918} & \textbf{0.7851} & \textbf{0.7556} & \textbf{0.7542} \\
w/o $h_{\text{dir}}$                                & 0.7202          & 0.7246          & 0.6196          & 0.5933          & 0.8149          & 0.8184          & 0.8087          & 0.8164          & 0.7345          & 0.7307         & 0.7778          & 0.7754          & 0.7460          & 0.7431          \\
w/o $h_{\text{ind}}$                                 & 0.7231          & 0.7281          & 0.6186          & 0.5945          & 0.8161          & 0.8161          & 0.8090          & 0.8129          & 0.7374          & 0.7374         & 0.7867          & 0.7775          & 0.7485          & 0.7444          \\
w/o $h$                           & 0.7215          & 0.7301          & 0.6136          & 0.5852          & 0.8142          & 0.8175          & 0.8124          & 0.8184          & 0.6889          & 0.6881         & 0.7719          & 0.7693          & 0.7371          & 0.7348          \\
w/o retrieval features                      & 0.7201          & 0.7217          & 0.6200          & 0.5806          & 0.8128          & 0.8137          & 0.8093          & 0.8132          & 0.7331          & 0.7282         & 0.7883          & 0.7719          & 0.7473          & 0.7382          \\
w/o fine-tuning features                    & 0.7126          & 0.7218          & 0.5916          & 0.5345          & 0.7299          & 0.7421          & 0.7496          & 0.7647          & 0.7364          & 0.7390         & 0.7387          & 0.7404          & 0.7098          & 0.7071          \\ \hline \hline
\end{tabular}
}
\end{table*}

\subsection{Implementation Details}
\label{subsec: implementation details}

Following prior studies~\cite{zhao2021missing, zuo2023exploiting, lian2023gcnet, liu2024contrastive, xu2024leveraging}, we evaluate our model under six missing-modality settings: \{a\}, \{t\}, \{v\}, \{a, t\}, \{a, v\}, and \{t, v\}, where ‘a’, ‘t’, and ‘v’ denote the acoustic, textual, and visual modalities, respectively. Each set indicates the modalities that remain available during inference.
To ensure fair comparison, we adopt publicly available features from~\cite{lian2023gcnet, xu2024leveraging}.
All models were trained for 25 epochs using the Adam optimizer with a learning rate of 0.0001 and a dropout rate of 0.5.
Hyperparameters were set as $k{=}5$, $\alpha_1{=}0.6$, $\alpha_2{=}0.4$, and $\beta{=}4$.
Experiments were conducted on NVIDIA A800 GPUs with PyTorch 1.13.1 and CUDA toolkit 11.1.1.

\subsection{Comparison with SOTA Methods}
\label{subsec: comparison with sota methods}

To evaluate the performance of our method under various missing modality conditions, we conduct comparisons with several state-of-the-art (SOTA) methods, including CPMNet \cite{zhang2020deep}, GCNet \cite{lian2023gcnet}, MMIN \cite{zhao2021missing}, CIF-MMIN \cite{liu2024contrastive}, and MoMKE \cite{xu2024leveraging}, on two benchmark datasets. All methods are tested under the same fixed missing modality settings.
As shown in Tab.~\ref{tab: main results}, our method consistently outperforms prior approaches in both per-condition and average performance across all testing conditions. Specifically, HARDY-MER achieves improvements of 0.0443, 0.0297, and 0.0143 in average WA on the IEMOCAP (4-class), IEMOCAP (6-class), and CMU-MOSEI tasks, respectively, demonstrating strong generalization and robustness under incomplete modality inputs.
In particular, we observe the most significant performance gain under the \{v\} condition. This may be attributed to the inherently higher uncertainty of visual features, which are more difficult to interpret in isolation. In such cases, our hardness-aware retrieval mechanism provides semantically relevant support samples, enhancing both representation quality and prediction reliability.
Although slight performance drops (approximately 0.9\% - 1.5\%) occur under the \{a,t\} and \{t,v\} conditions in CMU-MOSEI, our method still delivers the best overall performance, achieving improvements of 0.0132 and 0.0163 in ACC and F1, respectively.
These results further validate the effectiveness and practical applicability of HARDY-MER for robust multimodal learning with missing inputs.

\subsection{Ablation Study}
\label{subsec: ablation study}
To thoroughly investigate the effectiveness of different modules in our model, we designed a series of ablation experiments and validated them on the IEMOCAP four-class task:

1) w/o $h_{\text{dir}}$ \& w/o $h_{\text{ind}}$: To evaluate the individual impact of each hardness component, we perform ablation studies by removing either the direct hardness (w/o $h_{\text{dir}}$) or indirect hardness (w/o $h_{\text{ind}}$) from the overall sample hardness computation.
In the w/o $h_{\text{dir}}$ setting, we exclude the direct hardness term and calculate sample hardness solely based on the indirect hardness.
Conversely, in the w/o $h_{\text{ind}}$ setting, we rely only on the direct hardness for the sample difficulty estimation.
As shown in Tab. \ref{tab: ablation study}, both of them lead to performance drops, confirming that each type of difficulty provides complementary value.
Notably, excluding $h_{\text{dir}}$ results in larger degradation, highlighting its stronger correlation with reconstruction difficulty.

2) w/o $h$: To evaluate the overall effectiveness of the proposed sample hardness mechanism, we conduct an ablation in which the hardness score is entirely removed from the retrieval process.
Instead of adaptively determining the number of retrieved samples based on each sample’s difficulty, we assign a fixed Top-$k$ number of support samples to all training instances, regardless of their reconstruction or semantic complexity. 
The performance degradation reported in Tab. \ref{tab: ablation study} indicates that adaptive retrieval based on sample difficulty yields more effective results than uniform sampling.

3) w/o retrieval features: To examine the effectiveness of retrieval-based curriculum learning, we remove the retrieval mechanism entirely and train the model using only the original training samples.
No additional support samples are retrieved during training.
The results in the row of \textit{w/o retrieval features} in Tab. \ref{tab: ablation study} indicate that solely using the original samples, without allocating additional samples for challenging cases during training, diminishes the model’s training efficacy. This observation also validates the effectiveness of our retrieval curriculum.

4) w/o fine-tuning features: To assess the importance of feature quality in the retrieval process, we replace the fine-tuned features used for building the retrieval index with publicly pretrained features from prior work.
The results in Tab. \ref{tab: ablation study} show that the model achieves significant improvements after using fine-tuned features, especially in the {t} condition, indicating that high-quality features are crucial for maintaining retrieval accuracy and model robustness.

5) Hyperparameter ablation: 
To assess the impact of the hyperparameters in Eq. \ref{eq: hardness} on model performance, we conduct ablation studies on $\alpha_1$, $\alpha_2$, and $\beta$. We report the average WA and UA scores across six missing modality scenarios, as shown in Tab. \ref{tab: hyperparameters}. The results indicate that increasing $\alpha_1$ generally enhances performance, suggesting that direct hardness plays a more critical role in assessing overall sample hardness. However, when $\alpha_1$ exceeds 0.6, the contribution of indirect hardness is overly suppressed, leading to a decline in performance. The parameter $\beta$ serves to normalize the hardness metric within the [0, 1] range; if set too high or too low, it disrupts sensitivity and undermines the model’s ability to dynamically adjust the K-value, ultimately affecting overall performance.

\subsection{Visualization Analysis}
\label{subsec: visualization analysis}
\begin{figure*}[!htbp]
    \centering
    \begin{minipage}{0.3\textwidth}
        \centering
        \includegraphics[width=\linewidth]{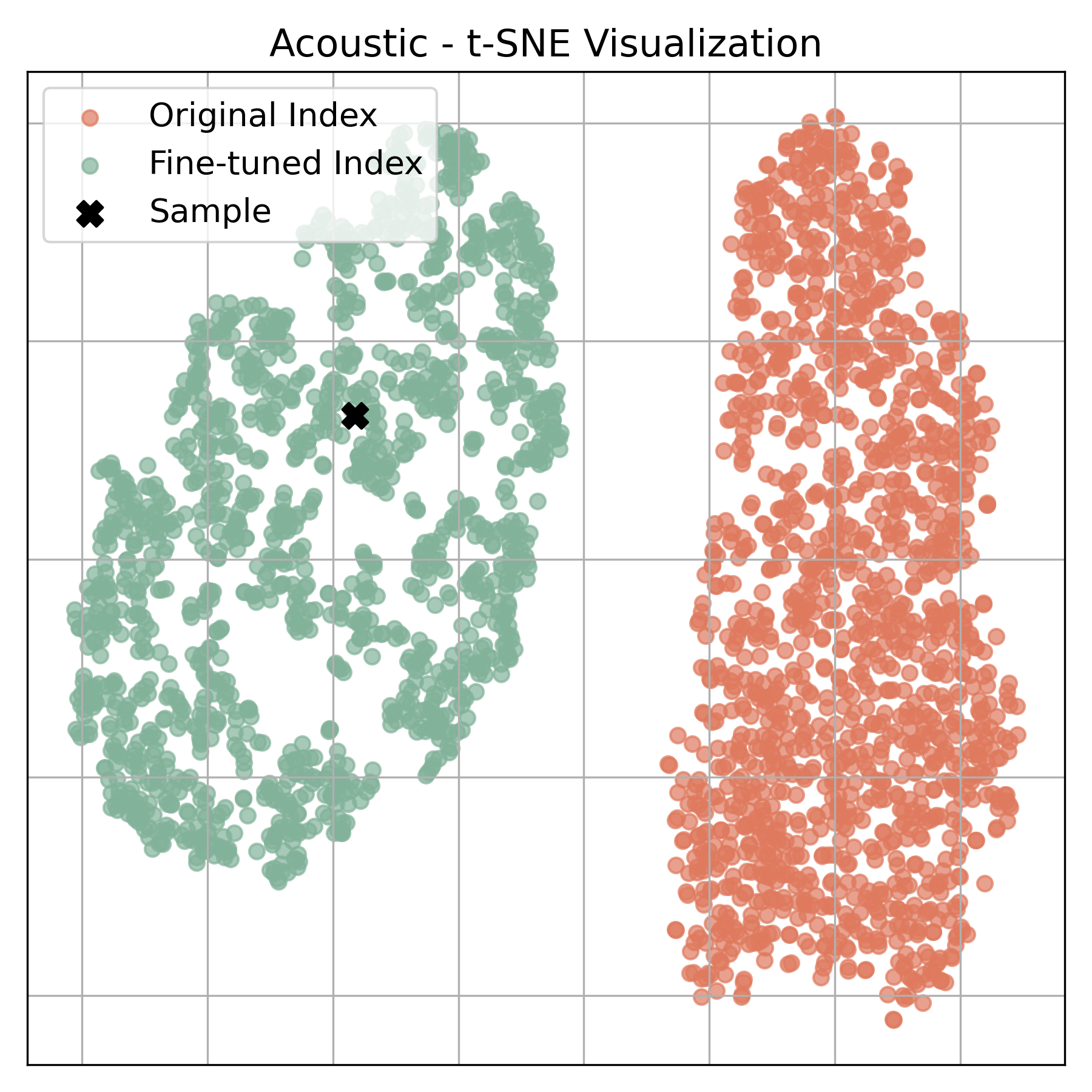}
    \end{minipage}\hfill
    \begin{minipage}{0.3\textwidth}
        \centering
        \includegraphics[width=\linewidth]{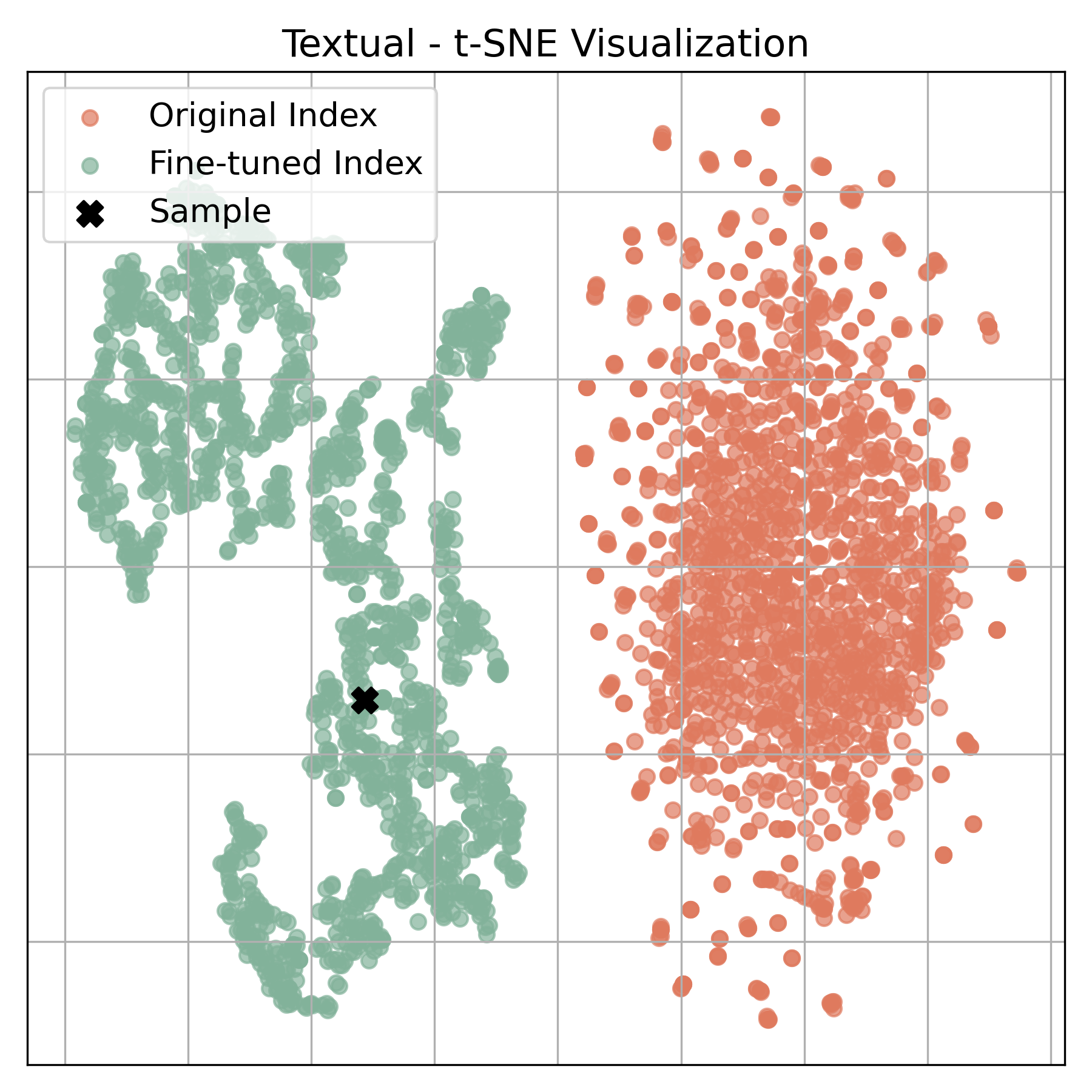}
    \end{minipage}\hfill
    \begin{minipage}{0.3\textwidth}
        \centering
        \includegraphics[width=\linewidth]{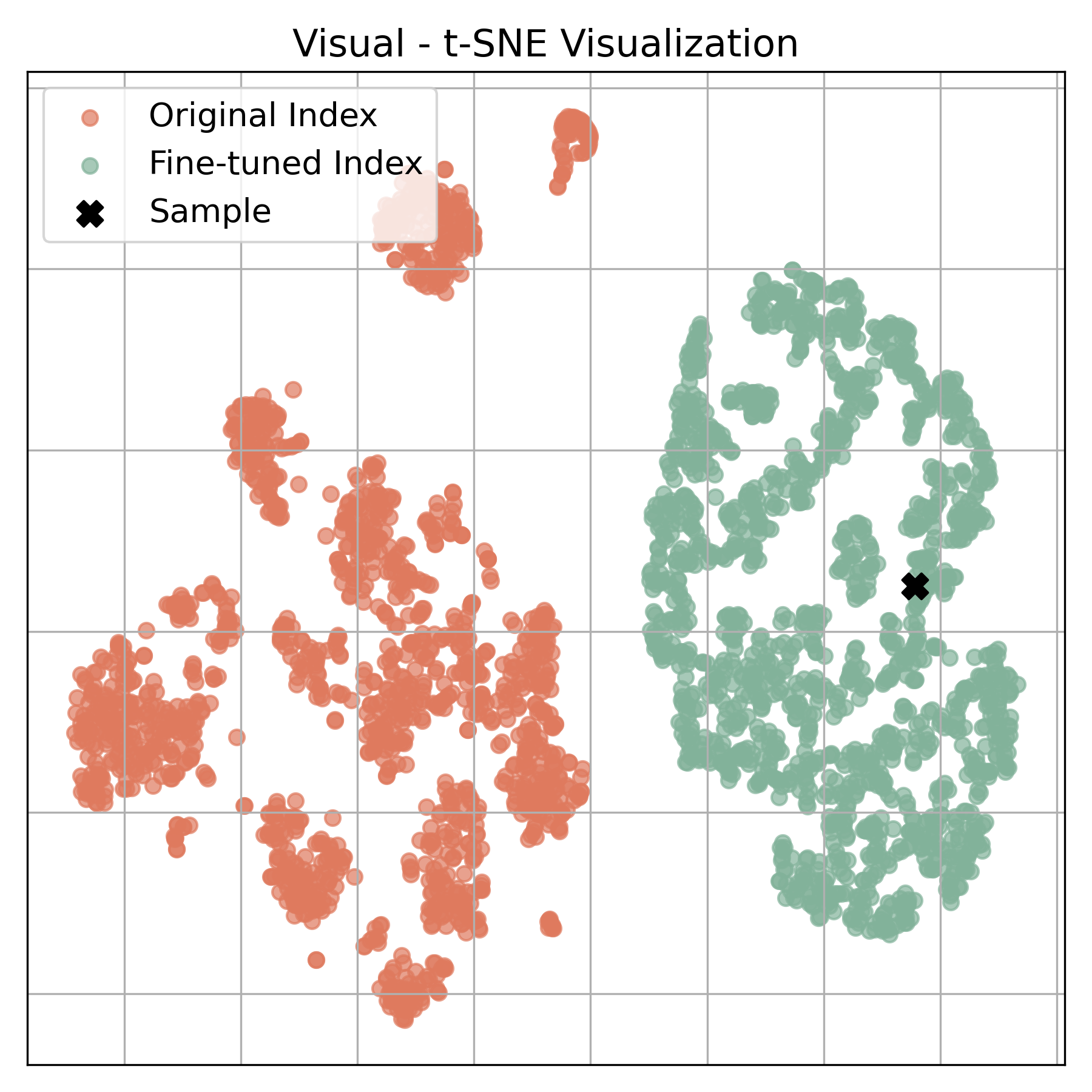}
    \end{minipage}
    \caption{t-SNE visualizations for randomly selected samples in the IEMOCAP four-class across acoustic, textual, and visual modalities.}
    \label{fig: fine-tune}
\end{figure*}

To analyze the impact of fine-tuning on similarity measurement, we visualized the retrieved samples using t-SNE.
We randomly selected a sample from the IEMOCAP dataset (four-class) and retrieved the top 1502 most similar samples to this sample from both the original feature index and the fine-tuned feature index.
Red and green points in Fig. \ref{fig: fine-tune} denote the original and fine-tuned features, respectively, while the black “X” marks the queried sample.
The results show that fine-tuned features are more concentrated around the queried point across all modalities, indicating improved retrieval accuracy after fine-tuning.

\begin{table}[!htbp]
\caption{The results of ablation study on the hyperparameters in Eq.~\ref{eq: hardness} on the IEMOCAP four-class task.}
\begin{tabular}{ccc}
\hline
\multirow{2}{*}{Setting} & \multicolumn{2}{c}{Average} \\ \cline{2-3} 
                         & WA           & UA           \\ \hline \hline
$\alpha_1=0.2$, $\alpha_2=0.8$, $\beta=4$              & 0.7500       & 0.7508       \\
$\alpha_1=0.4$, $\alpha_2=0.6$, $\beta=4$              & 0.7508       & 0.7517       \\
$\alpha_1=0.8$, $\alpha_2=0.2$, $\beta=4$              & 0.7512       & 0.7520       \\ \hline
$\alpha_1=0.6$, $\alpha_2=0.4$, $\beta=2$                   & 0.7512       & 0.7501       \\
$\alpha_1=0.6$, $\alpha_2=0.4$, $\beta=8$                   & 0.7550       & 0.7518       \\ \hline
our($\alpha_1=0.6$, $\alpha_2=0.4$, $\beta=4$)                      & 0.7556       & 0.7542       \\ \hline \hline
\end{tabular}
\label{tab: hyperparameters}
\end{table}

\begin{figure}[!htbp]
    \centering
    \begin{minipage}{\linewidth}
        \centering
        \includegraphics[width=\linewidth]{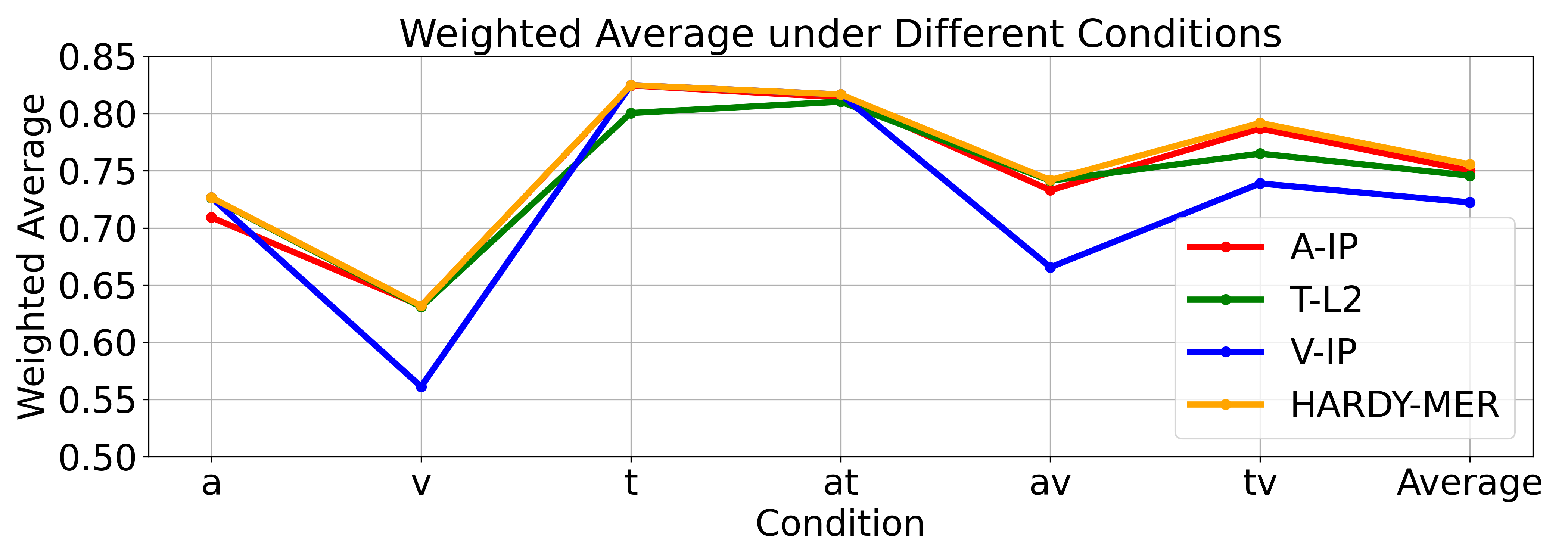}
    \end{minipage}\hfill
    \begin{minipage}{\linewidth}
        \centering
        \includegraphics[width=\linewidth]{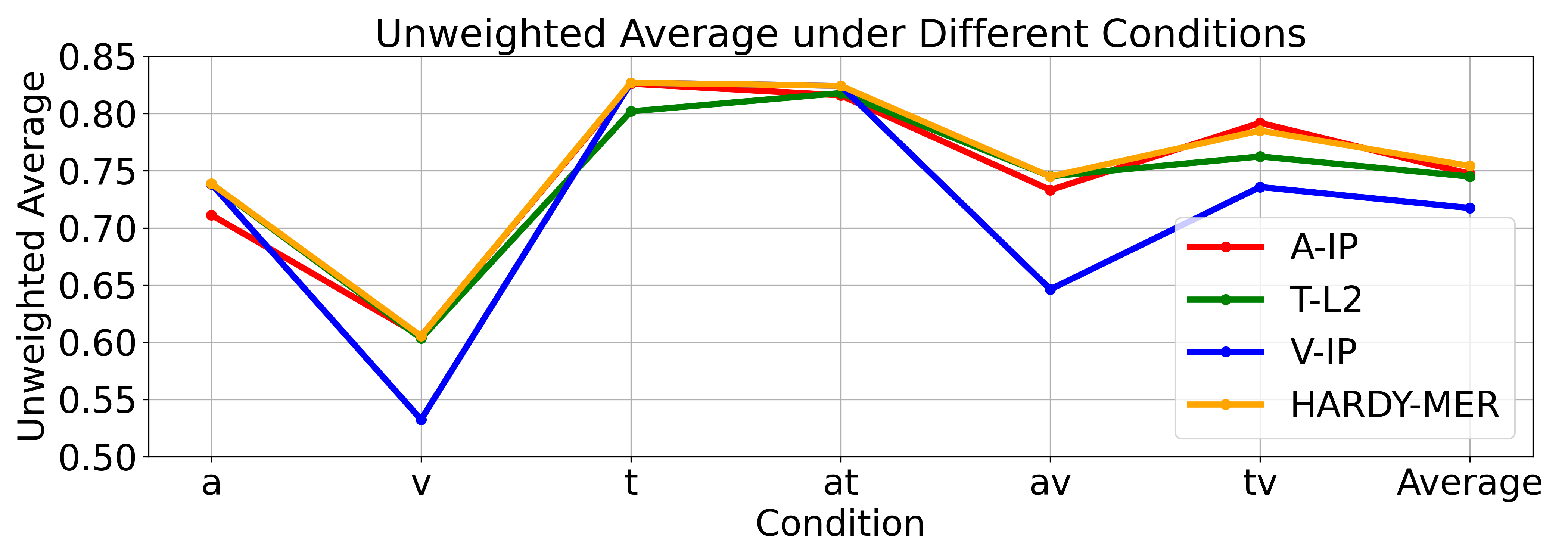}
    \end{minipage}\hfill
    \caption{Impact of different index construction methods on model performance, evaluated on the IEMOCAP (four-class) task. The line chart shows the variation of WA and UA across six missing modality conditions and their average.}
    \label{fig: index}
\end{figure}

We further investigate the impact of different index construction strategies by comparing our default setting with three alternatives, each modifying the distance metric of a single modality: 
1) \textbf{A-IP}: replaces \textit{IndexFlatL2} with \textit{IndexFlatIP} for the acoustic index; 
2) \textbf{V-IP}: applies \textit{IndexFlatIP} to the visual index; 
3) \textbf{T-L2}: uses \textit{IndexFlatL2} for the text index instead of \textit{IndexFlatIP}.
Fig.~\ref{fig: index} reports the Weighted Accuracy (WA) and Unweighted Accuracy (UA) under various modality conditions. Results show that using L2 distance for the text index (\textbf{T-L2}) consistently degrades performance, especially in text-only or text-involved settings (e.g., \textit{t}, \textit{at}, \textit{tv}), highlighting the suitability of inner product for normalized textual embeddings. In contrast, switching to cosine similarity for acoustic (\textbf{A-IP}) or visual (\textbf{V-IP}) indexing reduces accuracy, with \textbf{V-IP} showing the most notable drop, particularly under visual-only input. These findings suggest that L2 distance is more effective for acoustic and visual features, which typically retain important magnitude information.

\section{Conclusion}
\label{sec: conclusion}

To improve sensitivity to hard samples and enhance robustness in missing-modality multimodal emotion recognition, we propose HARDY-MER, a novel framework that combines retrieval-augmented learning with curriculum learning. 
We introduce a multi-view hardness evaluation mechanism based on reconstruction errors and cross-modal mutual information, and design a Retrieval-based Dynamic Curriculum Learning strategy. This involves retrieving semantically relevant support samples from modality-specific feature banks, with retrieval quantity adaptively determined by sample hardness.
The resulting hardness-aware curriculum guides model training.
Experiments show HARDY-MER outperforms state-of-the-art methods, and to our knowledge, it is the first to integrate retrieval and curriculum learning in this setting.
Future work will explore extending HARDY-MER to large-scale pre-trained multimodal models for greater robustness under challenging conditions.


\section{Acknowledgments}
The research by Rui Liu was funded by the Young Scientists Fund (No.~62206136), the General Program (No.~62476146) of the National Natural Science Foundation of China,  the Young Elite Scientists Sponsorship Program by CAST (2024QNRC001), the Outstanding Youth Project of Inner Mongolia Natural Science Foundation (2025JQ011), Key R\&D and Achievement Transformation Program of Inner Mongolia Autonomous Region (2025YFHH0014) and the Central Government Fund for Promoting Local Scientific and Technological Development (2025ZY0143). Zheng Lian was funded by the Excellent Youth Program of State Key Laboratory of Multimodal Artificial Intelligence Systems (No. MAIS2024311) and  Youth Science Fund Project of National Natural Science Foundation of China (No. 62201572).


\bibliographystyle{ACM-Reference-Format}
\balance
\bibliography{sample-base}

\end{document}